\documentclass{llncs}

\usepackage{graphicx}
\usepackage{textcomp}
\usepackage{listings}
\usepackage{subfiles}
\usepackage{caption} 
\usepackage{subcaption}
\usepackage{amsmath}
    \usepackage{wrapfig}
\usepackage{url}
\usepackage[belowskip=-5pt,aboveskip=0pt]{caption}
 
\begin{document}

\pagestyle{headings}  

\title{Survey on Semantic Stereo Matching / 
Semantic Depth Estimation
}
\titlerunning{short title}  
\author{Viny Saajan Victor\inst{1} \and Peter Neigel\inst{2}}
\authorrunning{Your name et al.} 
\institute{\email{victor@rhrk.uni-kl.de}
\and
\email{peter.neigel@dfki.de}}

\maketitle              

\begin{abstract}
Stereo matching is one of the widely used techniques for inferring depth from stereo images owing to its robustness and speed. It 
has become one of the major topics of research since  it finds its applications in 
autonomous driving, robotic navigation, 3D reconstruction, and many other fields. Finding pixel correspondences in non-textured, occluded and reflective areas is the major challenge in stereo matching. Recent developments have shown that semantic cues from image segmentation can be used to improve the results of stereo matching. Many deep neural network architectures have been proposed to leverage the advantages of semantic segmentation in stereo matching. This paper aims to give a comparison among the state of art networks both in terms of accuracy and in terms of speed which are of higher importance in real-time applications.

\end{abstract}

\section{Introduction}

\textbf{1.1 Stereo Matching:} Stereo Matching is a technique used to find point correspondences between two images of a scene acquired by known cameras. The traditional approaches such as 
    Semi-Global Matching \cite{SGM} includes well-defined steps such as cost matching (calculating matching cost for each pixel in one of the stereo images depending on the corresponding pixel in the other image),  cost aggregation (calculating matching cost for range of disparity values to form cost volumes), disparity estimation (calculating the disparity from cost volume by minimizing the cost function) and disparity post-processing (improving the disparity). These approaches make assumptions that the scene is uniformly illuminated and textured.\\
    \\
\textbf{1.2 Semantic Segmentation:} Semantic Segmentation is an image 
segmentation methodology that approaches the problem by performing pixel-level classifications.
Traditional approaches include superpixel segmentation \cite{Superpixel} (dividing an image into hundreds of non-overlapping superpixels), active contour methods \cite{ActiveContour} (methods which make use of the energy constraints and forces in the image for separation of region of interest) and watershed segmentation methods \cite{WaterShed} (algorithm which treats pixels values as a local topography).\\
\\
\textbf{1.3 Synergies between Stereo Matching and Semantic Segmentation:} Recent works in semantic matching \cite{SegStereo} have shown that the semantic labels obtained from semantic segmentation improve the accuracy of stereo matching in non-textured, occluded and reflective regions. Similarly, depth information obtained by stereo matching can be used to solve possible confusions between similar semantic categories \cite{SemForDipsEstimation}. This shows that segmentation information and depth maps complement each other and together they represent high-level information of the scene.\\
\\
\textbf{1.4 State of the Art:}
Deep learning techniques have shown great success in both stereo matching and semantic segmentation during the past ten years. The state of the art uses deep neural networks that are trained for one of these tasks exclusively and they achieve very reliable accuracy. Recent developments \cite{SegStereo}, \cite{SemForDipsEstimation} in the field revealed that stereo matching and semantic segmentation are intrinsically related, and both pieces of
information need to be considered in an integrated manner to succeed in challenging applications such as robotics and autonomous navigation. The interactive environment in such applications requires stereo matching to have a high-level understanding of the scene. Single deep neural networks jointly trained for both the tasks achieve better accuracy when compared to standalone models but are not efficient in terms of speed due to the network complexity. The work presented in this paper is aimed towards the possible exploration of various methods used in semantic
stereo matching for real-time applications and compare them in terms of accuracy (network loss) and speed (fps) on benchmark datasets. We also try to compare, to what extent these methods are utilizing the synergies between depth information and semantic cues for the benefit of the applications.

\section{Semantic Depth Estimation}

Semantic Depth Estimation is the method used to improve the depth estimation obtained from stereo matching using semantic cues(class lael for each pixel in an image).
\subsection{Classification}
We can majorly classify sematic depth estimation based on two criteria.\\
\\
Based on training data availability:\\
\textbf{Supervised Semantic Depth Estimation:}
These approaches require datasets with stereo images and disparity ground truth \cite{RTSS} which are used to train the network. The network loss is reduced by comparing the output of the network with the ground truth.\\
\\
\textbf{Unsupervised Semantic Depth Estimation:}
These approaches do not require disparity ground truth \cite{SemForDipsEstimation}, as they mainly rely on warping error. After calculating the depth and thereby disparity, the right stereo image is warped with the calculated disparity
and compared with the left stereo image. The network tries to reduce the error as the visual difference between both images. These approaches still use supervised learning for semantic segmentation.\\
\\
Based on network model used:\\
\textbf{Standalone Approach:}
In this approach, networks are trained for depth estimation and semantic segmentation exclusively and then the result of semantic segmentation is used to improve disparity.\\
\\
\textbf{Joint-Learning Approach:}
These approaches use a common architecture for both stereo matching and semantic segmentation in the initial stages to extract features that are generic for both the tasks and refine the outputs in the later stages.
\begin{figure}[!ht]
\centering
\includegraphics[scale = 0.4]{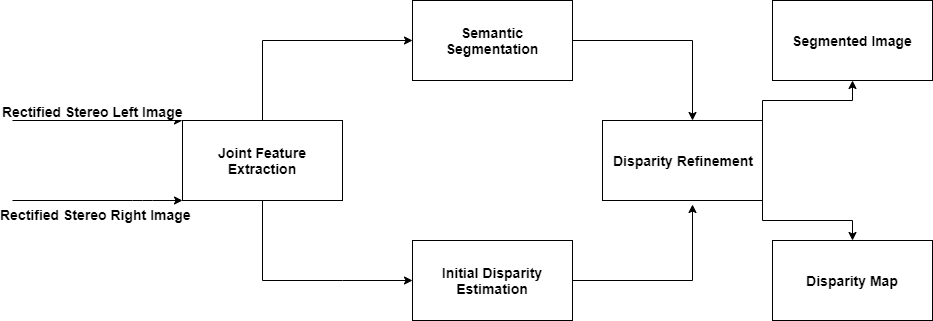}
\caption*{Figure 1: Stages involved in semantic depth estimation. Generic features are extracted from the input images in the joint feature extraction stage. The output of this stage is fed into 2 networks responsible for
    disparity estimation and semantic segmentation. The initial disparity obtained is improved using semantic cues in disparity refinement stage.}
\vspace{-3mm}
\end{figure}
\subsection{Stages in Extracting Depth from Stereo Images using Semantic Cues}
Depth from stereo images using semantic cues can be extracted in 4 generic stages.\\
\\
\textbf{2.2.1 Joint Feature Extraction Stage}\\
This stage extracts generic features from the input stereo images for stereo matching and semantic segmentation. Features for stereo matching are obtained in earlier layers and features for semantic segmentation are obtained from deeper layers since the latter requires more contextual information. The resolution of the images is usually reduced in this stage, extracting only relevant features for the task. The output of this stage is given to two separate subnetworks, one for depth estimation and the other for semantic segmentation. Hence the network learns a generic representation meaningful for both the tasks. Dovesi et. al\cite{RTSS} showed that this joint learning is efficient since a small number of features are sufficient for descent disparity estimation which in turn significantly increases the frame rate.
In \cite{RTSS}, the number of extracted features is given as hyper-parameter which provides flexibility  for the user to choose it according to the requirement (framerate vs accuracy).
\\
\\
\textbf{2.2.2 Disparity Estimation Stage}\\
This is the stage where the initial disparity is estimated. The feature map obtained from the joint feature extraction stage passes through deep convolution  layers to extract features that are specific for disparity estimation. Cost volumes are created for a range of disparity values and get the initial disparity map by applying regression. The disparity map obtained in this stage would contain error in non-textured, occluded and reflective areas of the images.\\
\textbf{Spatial Pyramid Cost Volumes for improving correspondences:}
\begin{figure}[!ht]
\begin{subfigure}[b]{0.5\textwidth}
\centering
\includegraphics[width= 6.5cm, height=4cm]{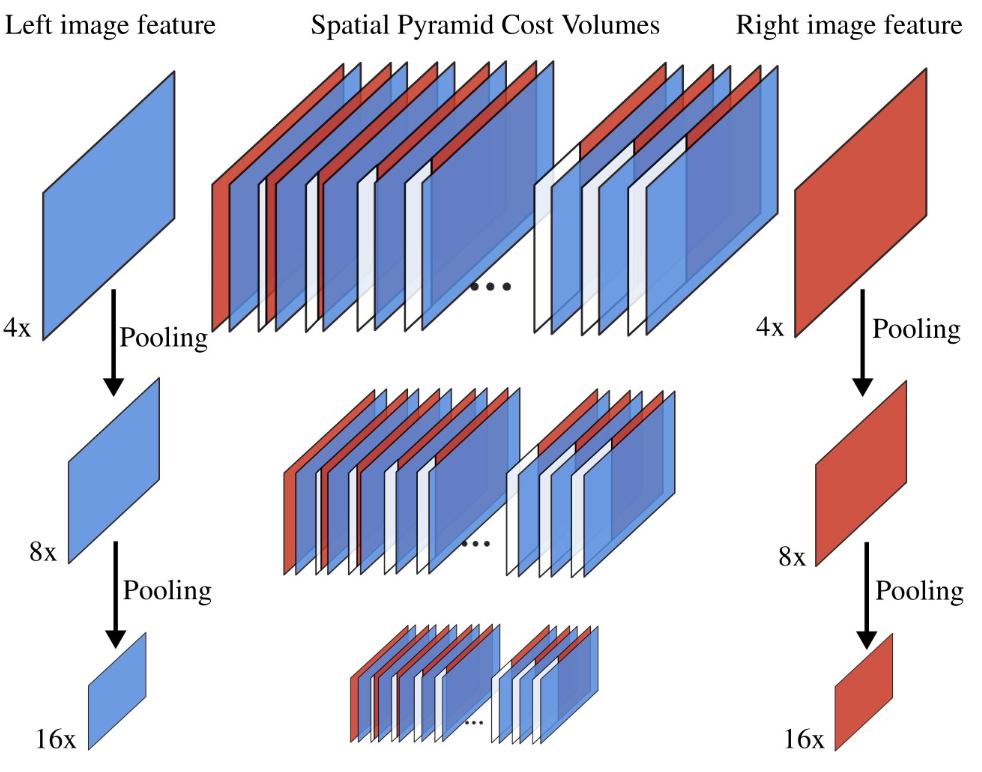}
\caption{}
\end{subfigure}%
\begin{subfigure}[b]{0.5\textwidth}
\centering
\includegraphics[width= 6.5cm, height=4cm]{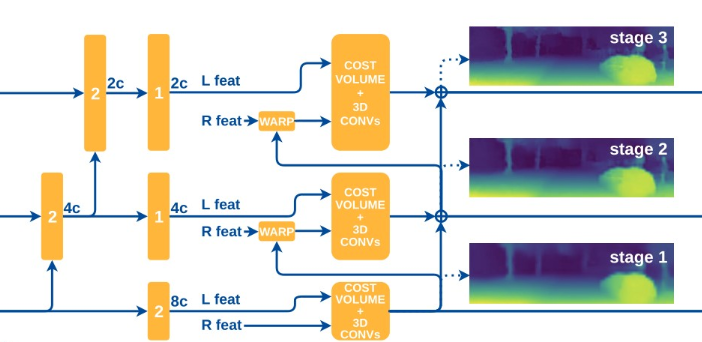}
\caption{}
\end{subfigure}%
\caption*{Figure 2(a): Construction of spatial pyramid cost volumes from left and right image features by spatial pooling. Figure is taken from \cite{SSPCV}.
Figure 2(b): coarse-to-fine disparity estimation. Each block contains the number of convolutional layers composing it and the number of features they output and C is the hyperparameter of the network which corresponds to the number of extracted features. Figure is taken from \cite{RTSS}.}
\vspace{3mm}
\end{figure}
Wu et. al\cite{SSPCV} used pyramidal cost volumes to learn the relationship between the object and its neighbors and thereby acquiring more correspondences. Figure 2(a) shows the construction process of spatial pyramid cost volumes from left and right image features by spatial pooling. For each resolution of feature maps, a 4D cost volume is formed by concatenating corresponding unaries from the left and right image features. The disparity can be inferred from this volume for that corresponding resolution.\\
\textbf{Optimizations to reduce the computation cost keeping desired accuracy:}
The network \cite{SemForDipsEstimation} concatenates every feature vector from the left image to all potential feature vectors from the right image which forms 5-dimensional cost volumes for both the views. The disparity information from these volumes is calculated by convolving it with a 3D filter which lets the network learn a better correlation metric during training. The memory intensity of 3D convolution is reduced by having an encoder-decoder architecture. In the network \cite{RTSS}, since the feature maps are extracted at different resolutions in the initial stage (1/16, 1/8, 1/4), this sub-network uses pyramidal network design where stack of decoders are used to decode the feature map from the joint feature extractor to estimate coarse-to-fine disparity maps as shown in Figure 2(b). This stage-wise disparity estimation at different resolution reduces computation cost as well as allowing the user to manage the speed-accuracy trade-off dynamically.\\
\\
\textbf{2.2.3 Semantic Segmentation Stage}\\
This stage is responsible for extracting semantic labels. Dovesi et. al\cite{RTSS} used the same coarse-to-fine estimation to achieve the benefits as explained in the previous stage.\\
\textbf{PSP Module \cite{PSP} for retaining contextual information from different scales:}
\cite{SemForDipsEstimation} has a pyramid parsing module for estimating semantic labels. This module obtains different sub-region representations which are upsampled and concatenated to form the final feature representation. By doing so both local and global context information can be carried to the later stage leading to better accuracy.\\
\\
\textbf{2.2.4 Disparity Refinement Stage}\\
Disparity computed in the initial disparity estimation stage contains noise and its' accuracy is reduced because of the poor matching in ill-posed regions. In this stage, disparity results will be refined using semantic cues based on the assumption that disparity values in the ill-posed regions will be similar to its neighbors because they belong to the same semantic segment (local smoothness constraint).
    The two embeddings (one from disparity subnetwork and the other from semantic segmentation subnetwork) are concatenated and this hybrid volume is passed through convolution layers to get the final refined disparity.\\
\textbf{Cost aggregation Module:}
Multiscale Context Intertwining \cite{intertwinning} is used to fuse the 4D spatial cost volumes from the lowest level to the higher ones in a recursive way. This module consists of two sub modules.\\
\begin{figure}[!ht]
\begin{subfigure}[b]{0.5\textwidth}
\centering
\includegraphics[width= 7.5cm, height=4cm]{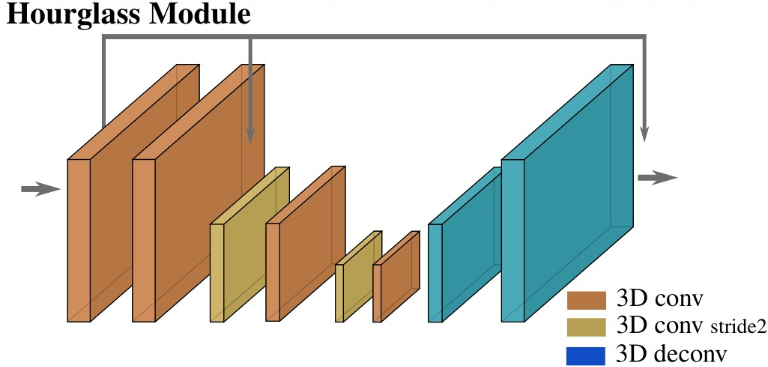}
\caption{}
\end{subfigure}%
\begin{subfigure}[b]{0.5\textwidth}
\centering
\includegraphics[width= 7.5cm, height=4cm]{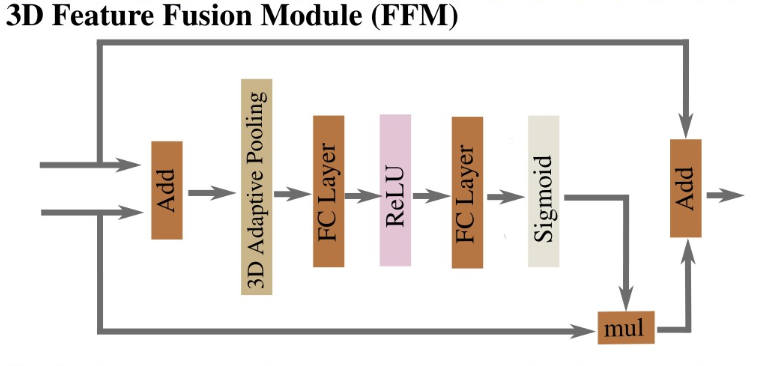}
\caption{}
\end{subfigure}%
\caption*{Figure 3: Components of 3D multi-cost aggregation module (a) Hour glass module is used to upsample the output and (b) 3D feature fusion module is used to concatenate features. Figure is taken from \cite{SSPCV}.}
\vspace{0mm}
\end{figure}
\textbf{i. Hour glass Module:}
To upsample the output Wu et. al\cite{SSPCV} used an hour-glass model that works on the principle of learnable upsampling. The paper shows that this approach is efficient because it retains more contextual information. As shown in figure 3(a), this encoder-decoder architecture consists of repeated top-down/bottom-up processing. Each module generates a disparity map for the corresponding resolution.\\
\textbf{ii. Feature Fusion Module:}
Instead of simply concatenating the features, a 3D feature fusion
module is used to retain the context information while fusing the cost volumes as shown in figure 3(b).\\
\textbf{Optimizations to reduce computational cost:}
During the refinement stage Zhang et. al\cite{SemForDipsEstimation} focused only on ill-posed regions using residual structures because mainly those regions are prone to error and thereby reducing computation cost.
\cite{RTSS} compresses the semantic embedding to have the same dimensionality of the disparity cost volume so that the fusion of these embeddings is computation effective.
\subsection{Loss Functions}
The overall loss function of the networks \cite{RTSS}, \cite{SemForDipsEstimation} and \cite{SSPCV} is given by
\begin{align}
L = \alpha L_{disp} + \beta L_{seg} + \gamma L_{ref}
\end{align}
where $L_{disp}$ is the initial disparity loss, $L_{seg}$ is the segmentation loss and $L_{ref}$ is the refined disparity loss,
\begin{table}[]
\centering
\begin{tabular}{|c|c|c|} \hline
\textbf{Loss} & \textbf{Stage} & \textbf{Learning type} \\ \hline
\cite{RTSS} L1- Smooth & Initial Disparity Estimation & Supervised \\ \hline
\cite{RTSS} Softmax Cross Entropy & Semantic Segmentation & Supervised \\ \hline
\cite{SemForDipsEstimation} Photometric & Initial Disparity Estimation & Unsupervised \\ \hline
\cite{SemForDipsEstimation} Regularization & Initial Disparity Estimation & Unsupervised \\ \hline
\cite{SemForDipsEstimation} Consistency & Initial Disparity Estimation & Unsupervised \\ \hline
\cite{SemForDipsEstimation} Smoothness  & Disparity Refinement & Supervised \\ \hline
\cite{SSPCV}Cross-domain discontinuity  & Disparity Refinement & Supervised \\ \hline
\end{tabular}
\caption{Losses used at different stages in the generic architecture shown in Figure 1 for Semantic Stereo Matching}
\label{tab:my-table}
\vspace{-5mm}
\end{table}
$\alpha$, $\beta$ and $\gamma$ are empirically determined constraints.\\
\textbf{2.3.1 L1-Smooth Loss \cite{huberloss} :}
Since disparity estimation from cost volume is a regression problem \cite{RTSS}, \cite{SSPCV} use Smooth L1-loss. This loss makes sure that the estimated disparity is as close as to the ground truth.\\
\begin{align}
L1_{smooth} = \begin{cases}0.5(d - \hat{d})^2 & \text{if $|d - \hat{d}|<1$;} \\
|d - \hat{d}| - 0.5 & \text otherwise\end{cases}
\end{align}
where d is the estimated disparity and $\hat{d}$ is the ground truth\\
Advantages of using L1-loss are that it has steady gradients for large values of error and it is less prone to oscillations during updates when the error is small.\\
\textbf{2.3.2 Softmax Cross Entropy \cite{cross} :}
Softmax Cross Entropy loss tries to reduce the divergence between the predicted probability of the semantic labels and the ground truth.
\begin{align}
L=-\sum_{c=1}^My_{o,c}\log(p_{o,c})
\end{align}
where,
$M$ is the total number of classes,
$y$ is the binary indicator if class label $c$ is the correct classification for observation $o$ and
$p$ is the predicted probability observation $o$ is of class $c$.\\
\textbf{2.3.3 Photometric Loss \cite{SemForDipsEstimation} :}
This loss works on the idea that the image reconstructed from estimated disparity should be very similar to the original image. It uses Euclidean distance(treats each feature equally) as well as structure similarity term SSIM $S$ ( metric that quantifies image quality degradation caused by processing such as data compression or by losses in data transmission) to improve the robustness in ill-posed regions. The photometric loss for the left image is
\begin{align}
L_p = \lambda_1S(I_L,I^{'}_L) + \lambda_2|I_L-I^{'}_L|+\lambda_3|\nabla I_L-\nabla I^{'}_L|
\end{align}
where $I_L$ and $I^{'}_L$ are left image and warped right image respectively, $\nabla$ is the gradient, $\lambda_1$,$\lambda_2$ and $\lambda_3$ are empirically found constants.\\
Since the loss depends on the warping error, it doesn't require a dataset with ground truth. But this loss introduces high frequency noise.\\
\textbf{2.3.4 Regularization Loss \cite{SemForDipsEstimation} :}
This loss is used to smooth local disparity from the input images based on the assumption that disparity in the local region tends to be smooth.
\begin{align}
L_r = \frac{1}{N}\sum|\nabla^2_xD_L|e^{-|\nabla^2_xI_L|} + |\nabla^2_yD_L|e^{-|\nabla^2_yI_L|}
\end{align}
where $D_L$ is the predicted left disparity, N is the number of pixels, $\nabla^2_x$ and $\nabla^2_y$ are second derivatives along x and y axes.\\
This loss reduces the noise introduced from the photometric loss.\\
\textbf{2.3.5 Consistency Loss \cite{SemForDipsEstimation} :}
This loss forces the right and left branches of the Siamese network to be consistent with each other.
The left image can be warped to the right view and warp it back to the left
view by using left disparity and right disparity. Consistency loss reduces the difference between this double warped image with the input image.
\begin{align}
L_c = |I_L-I^{''}_L|+|I_R-I^{''}_R|
\end{align}
where $I^{''}_L$ and $I^{''}_R$ are double warped left and right images respectively.\\
\textbf{2.3.6 Smoothness Loss \cite{SemForDipsEstimation} :}
This loss is used in the refinement stage where the disparity is improved using semantic cues. This is based on the idea that disparity should be smooth within a segment. This is conditioned on relatively good disparity.
\begin{gather}
L_s=\sum|\nabla^2_xD_L|(e^{-|\nabla^2_xf_L|}+e^{|Diff -t|})+\nabla^2_yD_L|(e^{-|\nabla^2_yf_L|}+e^{|Diff -t|})\\
Diff= \begin{cases}||D_L -D^{'}_L|| & \text{if $||D_L -D^{'}_L|| \leq t$}\\
t & \text otherwise\end{cases}
\end{gather}
where $f_L$ is the feature vector from the left view and t is an empirically found threshold.\\
\textbf{2.3.7 Cross-domain discontinuity loss \cite{Loss} :}
This loss is mainly aimed at enforcing an explicit link between the two learning tasks by leveraging on the ground truth pixel-wise semantic labels to improve depth prediction based on the assumption that depth discontinuities are likely to co-occur with semantic boundaries.
\begin{align}
L_{bdry}=\frac{1}{N}\sum_{i,j}(|\varphi_x(sem_{i,j})|e^{-|\varphi_x(d^{'}_{i,j})|}+|\varphi_y(sem_{i,j})|e^{-|\varphi_y(d^{'}_{i,j})|})
\end{align}
where $sem$ is semantic segmentation ground truth, $\varphi_x$ and $\varphi_y$ are the intensity gradients between neighboring pixels along the x and y directions, respectively and $d^{'}_{i,j}$ is the estimated disparity between pixels i and j. 
\section{Evaluation}
\textbf{Dataset:}
KITTI Vision Benchmark Suite 2015 \cite{KITTI} is used as the dataset for the evaluation which consists of 200 training scenes and 200 test scenes. It also comprises dynamic scenes for which the ground truth has been established in a semi-automatic process. It considers a pixel to be correctly estimated if the disparity or flow end-point error is lesser than 3 pixels. \\
\textbf{Metrics used for Evaluation:}
D1 is the percentage of stereo disparity outliers in the first frame,
all indicate percentage of outliers averaged over all ground truth pixels,
bg indicates the percentage of outliers averaged only over background regions and fg indicates the percentage of outliers averaged only over foreground regions.\\
\textbf{3.1 Average test set performance:}
From Table 2, we can infer that \cite{SSPCV} gives the best accuracy on the dataset but it is slower because of the heavy architecture. \cite{RTSS} gives a real-time performance with comparable accuracy.
Accuracy of \cite{SemForDipsEstimation} is less when compared to the other two networks.\\
\begin{table}[]
\parbox{.45\linewidth}{
\centering
\begin{tabular}{|c | c | c | c| c|} \hline
{\bf Network}  & {\bf D1-all} & {\bf Seconds/Frame}\\ \hline
RT$S^2$  & 3.56 \% & 0.02 \\ 
DispSegNet  & 6.33 \% & 0.9\\
SSPCV  & 1.85 \% & 0.9\\ \hline
\end{tabular}
\caption{Shows the average test set performance  on KITTI-2015 dataset \cite{KITTI}.}
\label{tab:my-table}
}
\hfill
\parbox{.45\linewidth}{
\centering
\begin{tabular}{|c | c | c | c | c | c | c | c|} \hline
{\bf $L^{init}_{p}$} & {\bf $L^{init}_{c}$} & {\bf $L^{init}_{r}$} & {\bf $L^{ref}_{p}$} & {\bf $L^{ref}_{c}$} & {\bf $L^{ref}_{s}$} & {\bf $L_{seg}$} & {D1-all\%}\\ \hline
Yes & Yes & Yes & No & No & No & No & 8.75\\
Yes & Yes & Yes & Yes & Yes & No & No & 8.60\\
Yes & Yes & Yes & Yes & Yes & Yes & No & 6.94\\
Yes & Yes & Yes & Yes & Yes & Yes & Yes & 6.32\\ \hline
\end{tabular}
\caption{Ablation study to show the impact of different losses used in \cite{SemForDipsEstimation} on accuracy.}
}
\vspace{-5mm}
\end{table}
\\
\textbf{3.2 Comparison between the methods in each stage}\\
Table 4 shows different modules/methods used in different stages of semantic depth estimation. From Tables 2 and 4 we can infer that at joint feature extraction stage, dilated convolutions are giving better accuracy when compared to encoder-decoder architecture because they preserve the resolution of the input and spatial pyramid cost volumes are contributing to the increase in  accuracy when compared to single spatial cost volumes beacause they learn the relationship between object and it's neighbours to improve correspondences. Even though hour-glass and 3dff modules used by \cite{SSPCV} at the disparity refinement stage increase the accuracy,  they are making the network slow because of their heavy fusion operations.\\
\begin{table}[]
\vspace{-3mm}
\centering
\begin{tabular}{|c | c | c | c | c |}  \hline
{\bf Network} & {\bf JointFeatExt} & {\bf DispCostVolume } & {\bf SegCostVolume} & {\bf DispRefine}\\ \hline
SSPCV & DilatedConvolution & Spatial Pyramid & Single Spatial & Hour-Glass and 3DFF\\
RTS2 & Encoder architecture & Spatial Pyramid & Spatial Pyramid & CascadeResidualConcatenate \\
DispSegNet & ResNet50(Siamese) & Single Spatial & Single Spatial & Concatenation\\ \hline
\end{tabular}
\caption{Different methods/modules used in semantic depth estimation stages}
\label{tab:my-table}
\vspace{-6mm}
\end{table}
\\
\textbf{3.3 Ablation Studies:}\\
In this section, we compare how much each module in the network is contributing to the overall objective.\\
\textbf{DispSegNet:} Table 3 shows the ablation study for different loss terms in \cite{SemForDipsEstimation}  where, $init$ is the initial disparity estimation, $ref$ is the disparity refinement, p is the photometric loss, c is consistency loss, r is regularization loss, s is smoothness loss and $seg$ is the supervised semantic segmentation loss. From the Table, we can observe that smoothness loss in the refinement stage is contributing significantly to the accuracy because they refine the disparity using semantic cues.\\
\textbf{SSPCV:} We can observe from Table 5 that pyramid cost volumes, feature fusion module are contributing to the accuracy beacause of context information retention and boundary loss function is predominantly contributing to the accuracy since it is aimed at enforcing an explicit link between the two learning tasks by leveraging on the ground truth pixel-wise semantic labels to improve depth prediction.\\
\textbf{RTS2:} We observe from Table 6 that stage 2 of \cite{RTSS} (i.e 1/8 resolution) is giving comparable accuracy with very good framerate.
\begin{table}
\centering
\begin{tabular}{|c | c | c | c | c | c |} \hline
{\bf SemBranch} & {\bf PyramidCost} & {\bf DilatedConv} & {\bf FFM} & {\bf BoundaryLoss} & {\bf D1-all\%}\\ \hline
Yes & No & No  & No & No & 2.37\\
Yes & Yes(3DMultiCost) & No & No & No & 1.99\\
Yes & Yes & Yes & No & No & 2.10\\
Yes & Yes & Yes & Yes & No & 1.93\\
Yes & Yes & Yes & Yes & Yes & 1.85\\ \hline
\end{tabular}
\caption{Ablation study to compare different model variants of \cite{SSPCV} to justify the selection criterion.}
\vspace{-5mm}
\end{table}
\begin{table}
\centering
\begin{tabular}{|c | c | c | c | c | c|} \hline
{\bf Init Disp Module} & {\bf Sem Seg Module} & {\bf Disp Refine Module} & {\bf Stage} & {\bf D1-all\%} & {\bf FPS}\\ \hline
Yes & Yes & Yes & 1 & 8.00 & 17.2  \\
Yes & Yes & Yes & 2 & 4.70 & 10.9\\
Yes & No & No & 3 & 3.98 & 8.1\\
Yes & Yes & No & 3 & 3.90 & 6.6\\
Yes & Yes & Yes & 3 & 3.33 & 6.3\\ \hline
\end{tabular}
\caption{Ablation study to show the impact of different modules of \cite{RTSS} on accuracy and framerate.}
\vspace{-10mm}
\end{table}
\section{Conclusion}
\vspace{-1mm}
From all the comparisons between the networks, we can infer that the networks which are focusing on accuracy are slower because of their heavy architecture. And networks focusing on framerate are compromising on accuracy. One has to always decide on the trade-off between time and accuracy depending on the application requirements. We also examined network which provides the flexibility for the user to choose between these two entities depending on their requirements. As future improvement, instance segmentation can be used for improving the depth estimation.
\vspace{-5mm}
%

%
%

%
\end{document}